\documentclass[letterpaper]{article} 
\usepackage[]{aaai23}  
\usepackage{times}  
\usepackage{helvet}  
\usepackage{courier}  
\usepackage[hyphens]{url}  
\usepackage{graphicx} 
\usepackage{natbib}  
\usepackage{caption} 
\usepackage{algorithm}
\usepackage{algorithmic}

\usepackage{newfloat}
\usepackage{listings}
\DeclareCaptionStyle{ruled}{labelfont=normalfont,labelsep=colon,strut=off} 
\floatstyle{ruled}
\newfloat{listing}{tb}{lst}{}
\floatname{listing}{Listing}
\title{Compressed Heterogeneous Graph for Abstractive Multi-Document Summarization}
\author {
    Miao Li, Jianzhong Qi, Jey Han Lau
}
\affiliations{
    School of Computing and Information Systems,\\
    The University of Melbourne\\
    miao4@student.unimelb.edu.au, \{jianzhong.qi, laujh\}@unimelb.edu.au
}

\usepackage{booktabs}       
\usepackage{amsfonts}       
\usepackage{nicefrac}       
\usepackage{graphicx}       
\usepackage{natbib}       
\usepackage{amsmath, amssymb, amsfonts, bm, mathtools}       
\usepackage{makecell}
\usepackage{multirow}
\usepackage{subfigure}
\usepackage{enumitem}
\usepackage{tabularx}
\usepackage{xspace}
\usepackage{adjustbox}

\newcommand{\tabref}[2][]{Table#1~\ref{tab:#2}}
\newcommand{\figref}[2][]{Figure#1~\ref{fig:#2}}

\usepackage[pdftex,dvipsnames]{xcolor}  
\definecolor{Mulberry}{rgb}{0.77,0.29,0.55}
\definecolor{CadmiumOrange}{rgb}{0.93,0.53, 0.18}
\definecolor{ForestGreen}{rgb}{0.13, 0.55, 0.13}
\definecolor{WildStrawberry}{rgb}{0.5, 0.7, 0.2}

\newcommand{\method}[1]{\textsc{#1}\xspace}
\newcommand{\hgsum}{\method{HGSum}}

\newcommand{\dataset}[1]{\textsc{#1}\xspace}
\newcommand{\multinews}{\dataset{Multi-News}}
\newcommand{\wcep}{\dataset{WCEP-100}}
\newcommand{\arxiv}{\dataset{Arxiv}}

\begin{document}

\maketitle

\begin{abstract}
Multi-document summarization (MDS) aims to generate a summary for a number of related documents.
We propose \hgsum\ --- an MDS model that extends an encoder-decoder architecture to incorporate a \emph{heterogeneous} graph to represent different semantic units (e.g., words and sentences) of the documents.
This contrasts with existing MDS models which do not consider different edge types of graphs and as such do not capture the diversity of relationships in the documents.
To preserve only key information and relationships of the documents in the heterogeneous graph,
\hgsum uses graph pooling to compress the input graph. And to guide \hgsum to learn the compression, we introduce an additional objective that maximizes the similarity between the compressed graph and the graph constructed from the ground-truth summary during training. \hgsum is trained end-to-end with the graph similarity and standard cross-entropy objectives.
Experimental results over \multinews, \wcep, and \arxiv show that \hgsum outperforms state-of-the-art MDS models.
The code for our model and experiments is available at: https://github.com/oaimli/HGSum.
\end{abstract}


\section{Introduction}

\emph{Multi-document summarization} (MDS) aims to automatically generate a concise and informative summary for a cluster of topically related source documents~\citep{mds_survey_2020, special_issues_2002}. It has a wide range of applications such as creating news digests~\citep{multinews_2019}, product review summaries~\citep{summarization_product_2014}, and summaries for scientific literature~\citep{damen_2022, summarization_biomedical_2022}. Our work targets \emph{abstractive} MDS, which generates summaries with words that do not necessarily come from the source documents, resembling the summarization process of human beings.



State-of-the-art text summarization models use \emph{pre-trained language models} (PLMs) including both general-purpose PLMs for text generation~\citep{led_2020, bart_2020} and PLMs designed for text summarization~\citep{pegasus_2020, primera_2022}. When applied to the abstractive MDS task, these models take a flat concatenation of the (multiple) source documents, which may not capture cross-document relationships such as contradiction, redundancy, or complementary information very well~\citep{cross_document_relationship_2004}. \citet{mds_survey_2020} argue that explicit modeling of cross-document relationships can potentially improve the quality of summaries. 
Following this, several recent studies~\citep{graphsum_2020, mgsum_2020, tg_multisum_2021} explore graphs to model source documents to improve abstractive MDS. However, these graphs are \emph{homogeneous} in that the nodes or edges are not distinguished for different semantic units (e.g.,\ words, sentences, and paragraphs) in the encoding process.
This means these MDS models cannot capture the diverse cross-document relationships among different types of semantic units.

\begin{figure}[t]
\centering
\includegraphics[width=0.35\textwidth]{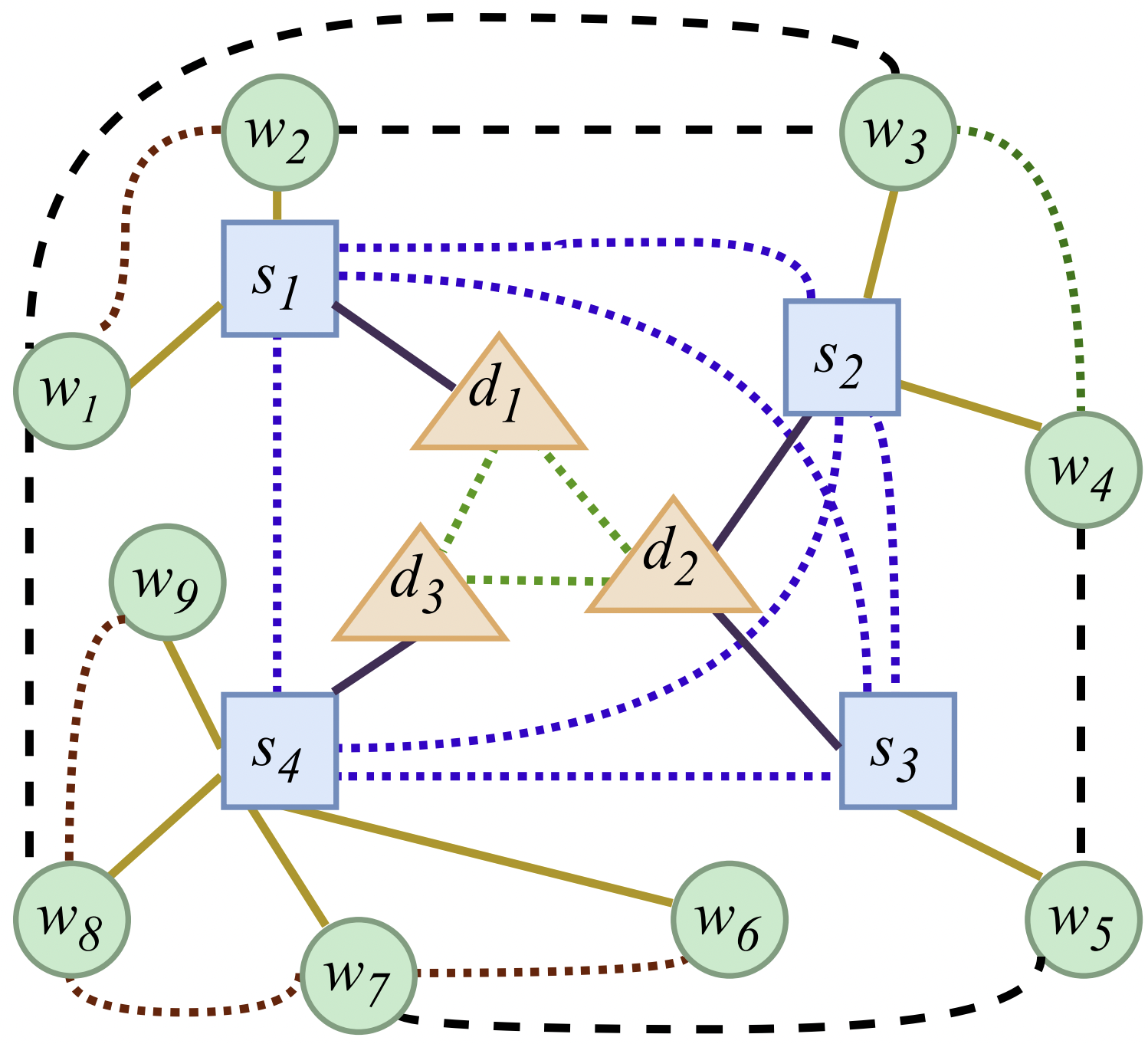}
\caption{The structure of the heterogeneous graph given three documents in a document cluster: The orange triangles denote document nodes $d$, the blue quadrates denote sentence nodes $s$, the green circles denote word nodes $w$, and the line (or curve) segments between nodes denote edges. A detailed description of the graph is in the Preliminaries. 
}
\label{fig:graph}
\end{figure}

In this paper, we propose \hgsum\ --- an MDS model that extends an encoder-decoder architecture to incorporate a heterogeneous graph to better capture the interaction between different semantic units in the documents.
\hgsum's heterogeneous graph has different types of nodes and edges to model words, sentences, and documents, as shown in~\figref{graph}.
To facilitate \hgsum to learn cross-document relationships, we construct edges between sentences \emph{across documents} based on the similarity of their sentence embeddings.
We also explore compressing the graph with graph pooling to preserve only salient information (i.e., nodes and edges) that is helpful for summarization, before feeding signals from the compressed graph to the text decoder to generate the final summary. To guide \hgsum to learn this compression, we introduce an auxiliary objective that maximizes the similarity between the compressed graph and the graph derived from the ground-truth summary, in addition to the standard cross-entropy objective during training.

There are several challenges that we face. First, it is non-trivial to encode heterogeneous graphs with existing graph neural networks,
as different types of nodes and edges should not be processed by the same function. To address this challenge, we propose multi-channel graph attention networks to encode heterogeneous graphs. Second, there are few graph compression or pooling methods proposed for heterogeneous graphs. Inspired by \citet{sag_pool_2019}, we introduce a compression method based on self-attentions to condense the heterogeneous graph. One novelty of our method is that it uses \textit{soft masking} so that 
it does not break the differentiability of the network, allowing us to train \hgsum in an end-to-end manner.

To summarize, our contributions are given as follows:

\begin{itemize}

\item We propose \hgsum, an MDS model that extends the  encoder-decoder architecture to incorporate a compressed graph to model the input documents. The graph is a heterogeneous graph that captures the diversity of semantic relationships in the documents, and it is compressed with a pooling method that helps preserve the most salient information for summarization.



\item \hgsum is trained with two objectives that maximize the likelihood of generating the ground-truth summary and the similarity between the compressed graph and the graph constructed from the ground-truth summary.

\item Experimental results over multiple datasets show that \hgsum outperforms state-of-the-art MDS models.
\end{itemize}

\begin{figure*}[ht]
\centering
\includegraphics[width=0.95\textwidth, trim=30 550 40 95, clip]{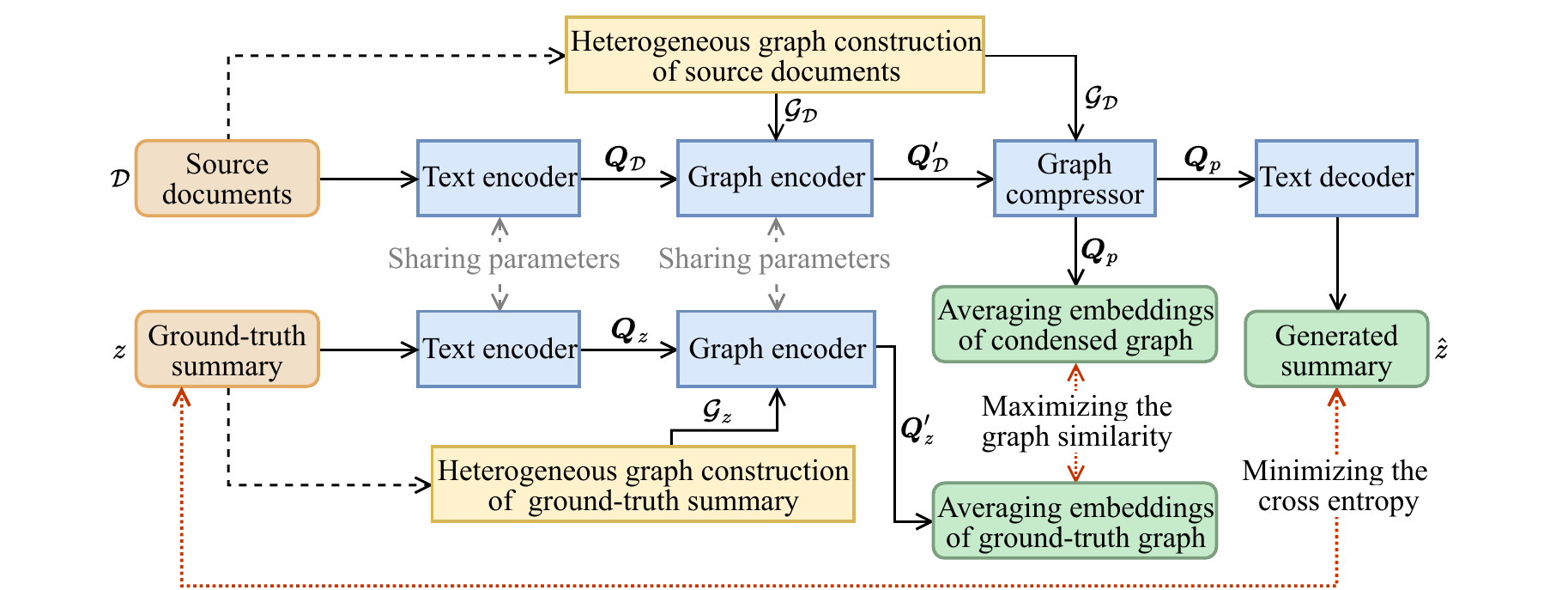}
\caption{The \hgsum architecture:  There are four main components: (1) {text encoder} (initialised using PRIMERA weights); (2) {graph encoder};  (3) {graph compressor}; and (4) {text decoder} (initialised using PRIMERA weights).
}
\label{fig:hgsum}
\end{figure*}

\section{Preliminaries}\label{sec:preliminaries}
Given a set of $m$ related source documents $\mathcal{D}=\{d_0, d_1, \ldots, d_m\}$ (i.e., a document cluster), our aim is to 
generate a text summary $\hat{z}=\hat{w_0}, \hat{w_1}, \ldots, \hat{w_T}$ (composed of $T$ words)
that captures the essence of the source documents.
As mentioned earlier, we generate the summary in an abstractive fashion, i.e., words in the generated summary can be words that are not found in the source documents.
The generation of each word in the summary is modeled as:
\begin{equation}
    \operatorname{p}(\hat{z}|\mathcal{D}) = \prod_{i=0}^T \operatorname{p}(\hat{w_i}|\mathcal{D}, \hat{w_0}, \hat{w_1}, \ldots, \hat{w_{i-1}})
\end{equation}


As heterogeneous graphs explicitly represent relationships among different semantic units (documents, sentences, and words), we construct a heterogeneous graph to represent a cluster of documents. 
We next explain how we construct the heterogeneous graph.



\subsection{Heterogeneous Graph Construction}
We denote the heterogeneous graph constructed to represent a cluster of documents as $\mathcal{G}=\langle\mathcal{V}, \mathcal{E}\rangle$, where $\mathcal{V}$ represents the set of nodes in the graph, and $\mathcal{E}$ the set of edges. As the example in~\figref{graph} shows, there are three types of nodes and six types of edges in $\mathcal{G}$. Specifically, $\mathcal{V}=\mathcal{V}_d\cup\mathcal{V}_s\cup\mathcal{V}_w$, where $\mathcal{V}_d$ is a set of document nodes: every document in the cluster corresponds to a node in $\mathcal{V}_d$ (orange triangles in~\figref{graph}); $\mathcal{V}_s$ is a set of sentence nodes: every sentence in the  documents corresponds to a node in $\mathcal{V}_s$ (blue quadrates in~\figref{graph}); and $\mathcal{V}_t$ is a set of word nodes\footnote{Technically, these are \emph{subword} nodes since we use subword tokenization, although most nodes map to full words in practice.}: every word 
in the sentences corresponds to a node in $\mathcal{V}_w$ (green circles in~\figref{graph}).

We next define the edges, which are all undirected:
\begin{itemize}
    \item 
    The sets $\mathcal{E}_{we}$ and $\mathcal{E}_{wo}$ contain edges between word nodes (dash and dot lines between word nodes in \figref{graph}). Every edge in $\mathcal{E}_{we}$ connects two nodes corresponding to noun words (identified based on a dependency parser\footnote{https://spacy.io/}).\footnote{Note that nodes that do not map to a full word will not have this type of edge, since they cannot be a noun.} The weight of an edge for a word pair in $\mathcal{E}_{we}$ is the cosine similarity of their embeddings. We use GloVe~\citep{glove_2014} as the static word embeddings in this work. Edges in  $\mathcal{E}_{wo}$, on the other hand, connect the nodes corresponding to every adjacent word pairs in a sentence.
    All edges in $\mathcal{E}_{wo}$ have a weight of 1.0.
    
    \item The set $\mathcal{E}_{ss}$ contains edges  that connect every pair of sentences (dot lines between sentence nodes in \figref{graph}). The weight of an edge for a pair of sentences is the cosine similarity of their pre-trained sentence embeddings. We use  Sentence-BERT~\citep{sentence_bert_2019} to compute the sentence embeddings, which is pre-trained based on the natural language inference task~\citep{snli_2015}. 
    
    
    \item The set $\mathcal{E}_{dd}$ contains edges  between document nodes (dot lines between document nodes in \figref{graph}). Every document is connected to all other documents in the cluster, and their edges are weighted using their n-gram overlap in terms of the average F1 value of ROUGE-1, ROUGE-2, and ROUGE-L \citep{rouge_2003}. 
    
    \item The sets $\mathcal{E}_{ds}$ and $\mathcal{E}_{st}$ 
    contain edges  that connect a document with its sentences (solid lines between document nodes and sentence nodes in \figref{graph}) and edges  that connect a sentence with its words (solid lines between sentence nodes and words nodes in \figref{graph}). These edges are designed to preserve the hierarchical document-sentence and sentence-word structures. All the edge weights in these sets are set to 1.0.
    
    \end{itemize}

To summarize, we have $\mathcal{E}=\mathcal{E}_{we}\cup\mathcal{E}_{wo}\cup\mathcal{E}_{ss}\cup\mathcal{E}_{dd}\cup\mathcal{E}_{ds}\cup\mathcal{E}_{sw}$. These edges collectively create a connected graph over all three types of nodes (words, sentences, and documents). Note that the choice of pre-trained word/sentence embeddings is flexible in our architecture, and in future work it would be interesting to explore other pre-trained embeddings.

\section{The \hgsum Model}

At its core, \hgsum extends a text encoder-decoder architecture (PRIMERA; \citet{primera_2022}) to incorporate information from a compressed heterogeneous graph derived from the input source documents, as presented in \figref{hgsum}. \hgsum has four main components: (1)~{text encoder} (initialized using PRIMERA weights), (2)~{graph encoder}, (3)~{graph compressor}, and (4)~{text decoder} (initialized using PRIMERA weights).

During training, we first generate two heterogeneous graphs $\mathcal{G}_\mathcal{D}$ and $\mathcal{G}_{z}$ based on the input source documents $\mathcal{D}$ and the ground-truth summary $z$, respectively, following the graph construction procedure described in the previous section. The text of input source documents $\mathcal{D}$ and ground-truth summary $z$ is processed by the text encoder to obtain contextual word embeddings $\boldsymbol{Q}_\mathcal{D}$ and $\boldsymbol{Q}_z$, respectively. These contextual word embeddings are then used by the graph encoder as the initial node embeddings of $\mathcal{G}_\mathcal{D}$ and $\mathcal{G}_z$, respectively. After processed by the graph encoder, we have the graph encodings $\boldsymbol{Q}'_\mathcal{D}$ and $\boldsymbol{Q}'_z$ respectively for the source documents and the ground-truth summary.\footnote{By graph encoding we mean the collective node embeddings in the graph.} The graph encoding of the source documents ($\boldsymbol{Q}'_\mathcal{D}$) will be further processed by the graph compressor to produce compressed graph encoding $\boldsymbol{Q}_p$, and this will be used by the text decoder to generate the final summary $\hat{z}$. To train \hgsum, we minimize the cross entropy between the ground-truth summary $z$ and generated summary $\hat{z}$ and maximize the similarity between the compressed graph encoding ($\boldsymbol{Q}_p$) and ground-truth summary graph encoding ($\boldsymbol{Q}'_{z}$).


Once the model is trained, we only use the text and graph encoders to encode the input source documents, the graph compressor to compress the document graph, and the text decoder to decode the summary, without using any ground-truth summary as input. We next detail these components.


\subsection{Text Encoder}

The text encoder follows the encoder architecture of PRIMERA --- which uses the sparse attention of longformer~\citep{led_2020} to accommodate long text input --- and is initialized with PRIMERA weights:
\begin{align}
    \boldsymbol{Q}_\mathcal{D} &= \operatorname{longformer}(\mathcal{D}) \\ \boldsymbol{Q}_z &= \operatorname{longformer}(z) 
\end{align}

The text encoder takes as input a concatenated string containing all the words from the documents, and it produces contextualized embeddings for these words as the output ($\boldsymbol{Q}_\mathcal{D}$ for source documents and  $\boldsymbol{Q}_z$ for the ground-truth summary). Note that we use special delimiters $\langle \textit{sent-sep}\rangle$ and $\langle \textit{doc-sep}\rangle$ to mark sentence and document boundaries, which allows us to extract sentence and document embeddings that we use as the initial sentence and document node embeddings in the graph encoder.



\subsection{Graph Encoder} 

The graph encoder is responsible for learning node embeddings for the document graph $\mathcal{G}_\mathcal{D}$ and the ground-truth summary graph $\mathcal{G}_{z}$. We explain how the graph encoder works for the document graph below, but the same principle works for processing the ground-truth summary graph.

Node embeddings for the heterogeneous graph $\mathcal{G}_\mathcal{D}$ represent the words, sentences, and documents, and they are initialized using the contextual embeddings learned from the text encoder ($\boldsymbol{Q}_\mathcal{D}$).
As standard graph neural networks (GNNs) based on message passing cannot be applied to the heterogeneous graphs directly, we propose \emph{multi-channel graph attention networks} (MGAT) inspired by graph attention networks (GAT; \citet{gat_2017}) to encode the heterogeneous graph. 

Similar to GAT, MGAT is a multi-layer graph network. Intuitively, in each layer, MGAT aggregates embeddings of different channels (i.e., edge types) for each node. 
The computation of the $l$-th layer of MGAT is given as follows:
\begin{align}
   \boldsymbol{h}_{i}^{(l+1)} &=\boldsymbol{U} \boldsymbol{H}^{(l)}_{i} \\
    \boldsymbol{H}^{(l)}_{i} &=\big\|_{c=1}^{C}\boldsymbol{h}_{i}^{(l), c}
\end{align}
where
$\boldsymbol{h}_{i}^{(l+1)}$ is the output embedding of node $i$ in the $l$-th layer, $\|$ is the concatenation operation, $C$ is the number of channels (which equals to the number of edge types in the heterogeneous graph, six in our case), and $\boldsymbol{U}$ is the shared transformation matrix for different nodes.
Intuitively, $\boldsymbol{h}_i^{(l), c}$ represents the embedding of node $i$ in the $c$-th channel at the $l$-th layer, and $\boldsymbol{H}^{(l)}_{i}$ is the concatenation of node embeddings from all channels for node $i$ in the $l$-th layer.
Note that the input node embeddings of the first layer of any channel 
are the output contextual embeddings (words, sentences, and documents) of the text encoder, i.e., $\boldsymbol{h}^{(0)}_i=\boldsymbol{q}_{i}$ where $\boldsymbol{q}_{i}\in \boldsymbol{Q}_\mathcal{D}$. The graph encoding, $\boldsymbol{Q}'_\mathcal{D}$, consists of all updated node embeddings from the final layer, i.e.,  $\boldsymbol{Q}'_\mathcal{D}=\big\|_{i}\boldsymbol{h}_{i}^{(L)}$.



To compute $\boldsymbol{h}_{i}^{(l), c}$ in each channel:
\begin{equation}
    \boldsymbol{h}_{i}^{(l), c}=\big\|_{m=1}^{M} \sigma\Big(\sum_{j \in \mathcal{N}_{i}^c} \alpha_{i j}^{m,c} \boldsymbol{W}^{m,c} \boldsymbol{h}_{j}^{(l), c}\Big) \\
\end{equation}
where $M$ is the number of attention heads. We can now see that $\boldsymbol{h}_{i}^{(l), c}$ is the concatenated representation of $M$ independent attention heads
with different weight matrices 
$\boldsymbol{W}^{m,c}$ and normalized attention weights  
$\alpha_{ij}^{m,c}$, with the latter computed as follows:
\begin{equation}
\label{eq:gat_attention_weight}
    \alpha_{i j}^{m,c=}\frac{\exp(d_{ij}^{m,c})}{\sum_{k \in \mathcal{N}_{i}^c} \exp(d_{ik}^{m,c})}
\end{equation}
where $\mathcal{N}_{i}^c$ denotes the set of nodes connected to node $i$ by an edge of type $c$. 
The attention coefficient $d_{ij}^{m,c}$ represents the correlation between nodes, and is learned as follows:
\begin{equation}
\label{eq:gat_attention_coefficient}
    d_{ij}^{m,c} = \sigma\big(e_{ij}\cdot\boldsymbol{w}_{m,c}^{\top} [\boldsymbol{W}^{m,c} \boldsymbol{h}_{i}^{(l),c} \| \boldsymbol{W}^{m,c} \boldsymbol{h}_{j}^{(l),c}]\big)
\end{equation}
where $e_{ij}$ is the  edge weight between node $i$ and node $j$ (defined in the Preliminaries section).


To summarize, MGAT computes node embeddings by attending to neighbouring nodes just like GAT, but it does this for each edge type independently and then concatenates  them together to produce the final node embeddings, and it repeats this for multiple layers/iterations to learn higher order connections. We note that
\hgsum has only one graph encoder, which is used to process both the source document graph $\mathcal{G}_\mathcal{D}$ to produce $\boldsymbol{Q}'_\mathcal{D}$ and the ground-truth summary graph $\mathcal{G}_{z}$ to produce $\boldsymbol{Q}'_z$.

\subsection{Graph Compressor} 


Given $\mathcal{G}_\mathcal{D}$ and $\boldsymbol{Q}'_\mathcal{D}$ from the graph encoder, the graph compressor aims to ``compress'' the graph by selecting a subset of salient nodes and edges. Here we focus on filtering the sentence nodes, because we want to identify key sentences that help generate the summary. After the compression, all selected sentence nodes \emph{and} their linked word and document nodes 
represent the compressed graph and their embeddings will be used by the text decoder for summary generation. 


The graph compressor is inspired by \citet{sag_pool_2019}, and it works by computing the attention scores for all sentence nodes, filtering out nodes with the lowest scores, and then masking the rest using their attention scores. Firstly, attention scores of the sentence nodes are calculated based on the updated node embeddings from our proposed graph encoder $\operatorname{MGAT}(\boldsymbol{Q}_\mathcal{D}, \mathcal{G}_\mathcal{D})$:
\begin{equation}
\label{eq:masking_score}
    \boldsymbol{t} = \operatorname{softmax}(\operatorname{MGAT}(\boldsymbol{Q}_\mathcal{D}, \mathcal{G}_\mathcal{D}) \cdot \boldsymbol{r})
\end{equation}
where $\boldsymbol{r}$ is the only trainable parameter of the graph compressor which transforms the updated node embedding into a scalar.
Then, based on these scores, we select sentence nodes with the highest scores:
\begin{align}
    \mathcal{I}_{s} &= \operatorname{top-k}(\boldsymbol{t}, k, \mathcal{G}_\mathcal{D}) \\
    \mathcal{I} &=\operatorname{extend}(\mathcal{I}_{s}, \mathcal{G}_\mathcal{D}) 
\end{align}
where $\operatorname{top-k}$ is a function that selects top-ranked sentence nodes in $\mathcal{G}_\mathcal{D}$ based on $\boldsymbol{t}$, $k\in (0, 1]$  is a hyper-parameter that determines the ratio of sentence nodes to be kept,  $\mathcal{I}_{s}$ is the set of selected sentence nodes, and  $\operatorname{extend}$ is a function that extends the selected sentence nodes in $\mathcal{I}_{s}$ to include word and document nodes that they link to (and so $\mathcal{I}$ includes word, sentence and document nodes).
Lastly, we mask all the selected nodes using their attention scores, producing the encoding of the compressed graph, $\boldsymbol{Q}_{p}$:
\begin{equation}
    \boldsymbol{Q}_{p}=\big\|_{i}^{\mathcal{I}}\boldsymbol{q}'_{i} \times \boldsymbol{t}_i, \boldsymbol{q}'_{i}\in \boldsymbol{Q}'_\mathcal{D}.
\end{equation}

\subsection{Text Decoder}

The text decoder follows the same architecture as a decoder Transformer (which uses masked attention to prevent attention to future words), is initialized with PRIMERA weights, and takes $\boldsymbol{Q}_p$ as input to generate the summary:
\begin{equation}
    \hat{z} = \operatorname{transformer}(\boldsymbol{Q}_p)
\end{equation}

Note that the node embeddings in $\boldsymbol{Q}_p$ retain the original word index in the source documents $\mathcal{D}$, and as such positional embeddings are added to them following standard transformer architecture.



\subsection{Multi-Task Training} 

\hgsum is trained with two objectives: maximizing the likelihood of generating the ground-truth summary $z$ and the graph similarity between the compressed graph encoding $\boldsymbol{Q}_p$ and ground-truth summary graph encoding $\boldsymbol{Q}'_z$.



To maximize the likelihood of generating the ground-truth summary, we minimize the cross entropy over the ground-truth summary and the generated summary with conventional teacher forcing. 
\begin{equation}
    \mathcal{L}_{ce}=-\frac{1}{T}\sum_{i=1}^T w_i\log \hat{w_i}
\end{equation}
where $w_i$ is the $i$-th word in the ground-truth summary, while $\hat{w_i}$ is the $i$-th word in the generated summary.
%

To maximize the graph similarity, we compute the cosine similarity of the average node embeddings from the compressed graph and the ground-truth summary graph:
\begin{equation}
    \mathcal{L}_{gs}=-\operatorname{sim}(\operatorname{avg}(\boldsymbol{Q}_p), \operatorname{avg}(\boldsymbol{Q}'_z))
\end{equation}

The final loss function of \hgsum is the sum of $\mathcal{L}_{ce}$ and $\mathcal{L}_{gs}$ weighted by hyper-parameter $\beta\in(0, 1)$. 
\begin{equation}
    \mathcal{L} = \beta\mathcal{L}_{ce} + (1-\beta) \mathcal{L}_{gs}
\end{equation}

\section{Experiments}
We test our proposed model \hgsum and compare it against state-of-the-art abstractive MDS models over several datasets. We also report the results of an ablation study to show the effectiveness of the components of \hgsum. 


\subsection{Experimental Setup}
\subsubsection{Datasets} We use \multinews~\citep{multinews_2019}, \wcep~\citep{wcep_2020},
and \arxiv~\citep{arxiv_2018} as benchmark English datasets. These datasets come from different domains including news, Wikipedia, and scientific domains.
\multinews contains clusters of news articles plus a summary corresponding to each cluster written by professional editors. \wcep contains human-written summaries of different news events from Wikipedia.
In \arxiv, each cluster corresponds to a research paper in the scientific domain, where the paper abstract is used as the summary, while sections of the paper are used as the source documents in each cluster.
\tabref{datasets} summarizes statistics of these datasets.


\begin{table}[t]
    \centering
    \setlength{\tabcolsep}{5pt}
    \begin{tabular}{rrrrr}
    \toprule
    \textbf{Dataset} & \textbf{\#c} & \textbf{\#d/c} & \textbf{\#w/d} & \textbf{\#w/s}\\
    \midrule
    \multinews & 56,216 & 2.79 & 690.97 & 241.61\\
    \wcep & 10,200 & 63.38 & 439.24 & 30.53\\
    \arxiv & 215,913 & 5.63 & 978.17 & 251.07 \\
    \bottomrule
    \end{tabular}
    \caption{Dataset statistics. ``c'' $=$ cluster; ``d'' $=$ document; ``w'' $=$ word; and ``s'' $=$ summary. ``$\#$'' denotes ``the number of'' and ``/'' denotes ``in each''.
    }
    \label{tab:datasets}
\end{table}

\begin{table}[t]
    \centering
    \begin{tabular}{rcrc}
    \toprule
    \textbf{Model} & \textbf{\#parameters} & \textbf{Len-in} & \textbf{Len-out}\\
    \midrule
    PEGASUS & 568M & 1,024 & 512\\
    LED & 459M & 16,384 & 512\\
    PRIMERA & 447M & 4,096 & 512\\
    MGSum & 129M & 2,000 & 400 \\
    GraphSum & 463M & 4,050 & 300 \\
    \hgsum & 501M & 4,096 & 512 \\
    \bottomrule
    \end{tabular}
    \caption{Model parameter sizes. Len-in and Len-out denote the maximum lengths of the model input and the model output, respectively.
    }
    \label{tab:model_settings}
\end{table}

\begin{table*}[t]
    \centering
    \begin{tabular}{rp{10mm}<{\centering}p{10mm}<{\centering}p{10mm}<{\centering}p{10mm}<{\centering}cp{10mm}<{\centering}p{10mm}<{\centering}p{10mm}<{\centering}p{10mm}<{\centering}}
    \toprule
    \multirow{2}{*}{\textbf{Model}} &  \multicolumn{3}{c}{\multinews} & \multicolumn{3}{c}{\wcep} & \multicolumn{3}{c}{\arxiv} \\
     & R-1 & R-2 & R-L & R-1 & R-2 & R-L & R-1 & R-2 & R-L\\
    \midrule
    PEGASUS & 47.70 & 18.36 & 43.62 & 42.43 & 17.33 & 32.35 & 44.21 & 16.95 & 38.87 \\
    LED & 47.68 & 19.72 & 43.83 & 43.05 & 20.94 & 34.99 & 46.50 & 18.96 & 41.87\\
    PRIMERA & \underline{49.40} & \underline{20.51} & \underline{45.35} & \underline{43.11} & \textbf{21.85} & \underline{35.89} & \underline{47.24} & \underline{20.24} & \underline{42.61}\\
    MGSum & 45.63 & 16.71 & 40.92 & 38.88 & 14.22 & 23.37 & 40.58 & 11.22 & 29.93\\
    GraphSum & 45.71 & 17.12 & 41.99 & 39.56 & 14.38 & 29.41 & 42.98 & 16.55 & 37.01 \\
    \hgsum (our model) & \textbf{50.64}\dag & \textbf{21.69}\dag & \textbf{45.90}\dag & \textbf{44.21}\dag & \underline{21.81} & \textbf{36.21}\dag & \textbf{49.32}\dag & \textbf{21.30}\dag & \textbf{44.50}\dag \\
    \midrule
     Performance gain & +2.51\% & +5.75\% & +1.21\% & +2.55\% & -0.18\% & +0.89\% & +4.40\% & +5.24\% & +4.44\% \\
    
    \bottomrule
    \end{tabular}
    \caption{Model performance on summarizing \multinews, \wcep, and \arxiv in terms of F1 of ROUGE scores. The best performance results are in boldface, while the second best is underlined. \dag: significantly better than others 
    ($\text{p-value}<0.05$).}
    \label{tab:mds_results}
\end{table*}

\subsubsection{Competitors}
We compare our model with two groups of state-of-the-art abstractive MDS models: \emph{PLM-based} and \emph{graph-based}.
(1) The PLM-based models include \textbf{PEGASUS}~\cite{pegasus_2020}, \textbf{LED}~\cite{led_2020}, and \textbf{PRIMERA}~\cite{primera_2022}.
LED is a general-purpose PLM that introduces the longformer architecture which uses sparse self-attention to allow it to process much longer input than previous models.
LED is pre-trained by reconstructing documents from their corrupted input in the same way as BART \citep{bart_2020}.
In contrast, PEGASUS and PRIMERA are pre-trained models designed for summarization (the former for single-document and the latter multi-document summarization).
Specifically, PEGASUS is pre-trained by generating pseudo summaries for documents, where the pseudo summaries are composed of gap sentences extracted from a document based on ROUGE scores. 
PRIMERA is similarly pre-trained to generate pseudo summaries, but their
 pseudo summaries are extracted based on the salience of entities which correspond to their document frequency.
For these PLM-based models, we take their off-the-shelf models and fine-tune them on our datasets. We follow the standard approach where we concatenate documents from the same cluster to form a long and flat input string.
(2) For the graph-based models, we compare against \textbf{MGSum}~\cite{mgsum_2020}\footnote{https://github.com/zhongxia96/MGSum} and \textbf{GraphSum}~\cite{graphsum_2020}\footnote{https://github.com/PaddlePaddle/Research/tree/master/NLP}.
To model cross-document relationships in MDS, MGSum~\citep{mgsum_2020} uses a three-level hierarchical graph to represent source documents, including different levels of nodes (documents, sentences, and words). It learns semantics with a multi-level interaction network. Although there are different types of nodes in this hierarchical graph, all of its edges are of the same type (i.e.,\ it is a homogeneous graph).\footnote{For fair comparison we use the abstractive variant of MGSum.}
GraphSum~\citep{graphsum_2020} uses a similarity graph over paragraphs to capture cross-document relationships, and it uses pre-trained RoBERTa~\citep{roberta_2019} as its encoder. Just like MGSum, its graph is homogeneous.


\subsubsection{Implementation Details} 

For the PLMs, we use the large version of the models which roughly have the same number of parameters (\tabref{model_settings}).\footnote{PLM-based models are implemented using the HuggingFace library: https://huggingface.co/}
For the graph-based models, we use open-source code from the original authors and train them on our datasets, following their recommended hyper-parameters and configurations.
As \tabref{model_settings} shows,
most models are trained to generate a maximum length of 512 subwords (``Len-out'') for the summary (exception: MGSum and GraphSum where we follow the original output length).
Note though that the maximum input lengths (``Len-in'') of these models range from 1K-16K subwords, depending on the architecture of their encoder. 

For \hgsum, the text encoder and decoder are initialized with PRIMERA weights.
To alleviate overfitting, we apply label smoothing during training with a smoothing factor of 0.1. We use beam search decoding with beam width 5 to generate the summary. The hyper-parameter $\beta$ is set to 0.5 to balance two loss functions.
All other hyper-parameters are tuned based on the development set.

All experiments are run on
Intel(R) Xeon(R) Gold 6326 CPU @ 2.90GHz with NVIDIA Tesla A100 GPU (40G).

\subsection{Overall Results}

We report the average F1 of ROUGE-1 (R-1), ROUGE-2 (R-2) and ROUGE-L (R-L)~\citep{rouge_2003}.
Note that we use the summary-level R-L,\footnote{We note that prior studies use a mixture of summary-level and sentence-level R-L, and for more details about their differences, we refer the reader to: https://pypi.org/project/rouge-score/} and each summary is split into sentences using NLTK\footnote{https://www.nltk.org/}. 

\tabref{mds_results} reports the performance of all models over all datasets. 
\hgsum  outperforms most of the benchmark systems, demonstrating the effectiveness of incorporating a compressed heterogeneous graph for text summarization. Interestingly, the PLMs (PEGASUS, LED, PRIMERA, and \hgsum) also seem to be consistently better than graph-based models (MGSum and GraphSum). This shows that using  graph-based document representations  does not necessarily lead to better MDS results, thus confirming the advantage of our  heterogeneous graph-based model design.
We give an example of generated summary by \hgsum in \multinews in \tabref{example_generated_summary}.



\begin{table}[t]
    \centering
    \begin{tabular}{p{15mm}<{}p{60mm}<{}}
    \toprule
    Doc 1 & $\ldots$ \textit{Parents are risking their babies' health because of a surge in the popularity of swaddling} $\ldots$\\
    Doc 2 & \textit{There has been a recent resurgence of swaddling because of} $\ldots$\\
    Doc 3 & $\ldots$ \textit{Swaddling babies is on the rise: Add it to the long list of mixed messages new parents get about infant care} $\ldots$\\
    \midrule
    Generated summary & \textit{The trend of swaddling babies is on the rise, but an orthopaedic surgeon $\ldots$ is warning parents against the practice.}\\
    \bottomrule
    \end{tabular}
    \caption{An example of a generated summary in \multinews by \hgsum.
    }
    \label{tab:example_generated_summary}
\end{table}

\subsection{Ablation Study}

To show the effectiveness of the \hgsum components, we conduct an ablation study and compare it with three model variants: 
(1)~\hgsum \textbf{w/o MGAT}, which replaces MGAT with the vanilla GAT model that treats all graph nodes and edges as being the same type, 
(2)~\hgsum \textbf{w/o graph compressor}, which drops the graph compressor from \hgsum and uses the output from the graph encoder directly as the input for the text decoder, and (3)~\hgsum \textbf{w/o multi-task training}, which replaces the multi-task objective using only the cross entropy objective.  

For the ablation results, we also present the performance in terms of BERTScore (``BScore''; \citet{bertscore_2020}), which measures the semantic similarity between the ground-truth and generated summary based on BERT embeddings.
\tabref{ablation_results} shows the ablation results on the test set of \multinews.\footnote{We found similar results for different datasets, and present only \multinews here in light of space.} We see that removing the heterogeneous graph encoder, graph compressor, or the multi-task objective result in a performance drop over all metrics, confirming the effectiveness of these components.
In particular,  dropping the multi-task objective leads to the largest degradation in model performance, suggesting that  this auxiliary task is essential to help \hgsum learn how to compress the graph for summarization.


\begin{table}[t]
    \centering
    \begin{tabular}{rp{6mm}<{\centering}p{6mm}<{\centering}p{7mm}<{\centering}p{10mm}<{\centering}}
    \toprule
    \textbf{Model} & \textbf{R-1} & \textbf{R-2} & \textbf{R-L} & \textbf{BScore}\\
    \midrule
    \hgsum & 50.64 & 21.69 & 45.90 & 87.38\\
    w/o MGAT & 48.87 & 20.32 & 43.21 & 87.08\\
    w/o graph compressor & 49.00 & 20.38 & 45.01 & 86.92\\
    w/o multi-task training & 48.10 & 20.30 & 44.24 & 86.85 \\
    \bottomrule
    \end{tabular}
    \caption{Results of ablation study on \multinews.
    }
    \label{tab:ablation_results}
\end{table}

\begin{table}[t]
    \centering
    \begin{tabular}{rcccc}
    \toprule
    \textbf{Initialized by} & \textbf{R-1} & \textbf{R-2} & \textbf{R-L} & \textbf{BScore}\\
    \midrule
        random weights & 18.99 & 27.86 & 16.88 & 79.32 \\
    LED & 48.36 & 19.99 & 44.25 & 86.73\\
    PRIMERA & 50.64 & 21.69 & 45.90 & 87.38\\
    \bottomrule
    \end{tabular}
    \caption{Summarization results of \hgsum with different initialization on  \multinews.}
    \label{tab:model_initialization}
\end{table}

\subsection{More Analysis}

\subsubsection{Impact of Text Encoder and Decoder Initialization} Our text encoder and decoder can be initialized by any pre-trained Transformer models. Here we make a comparison on initialization using PRIMERA, the large version of LED and random weights.  \tabref{model_initialization} shows results using such initialization strategies on the test set of \multinews.  
We see that initialization with random weights has much worse performance than initialization using pre-trained PLMs, which is expected. Using PRIMERA leads to better empirical performance than using the LED, consistent with prior findings.


\subsubsection{Impact of the Graph Compression Ratio $k$}
\begin{figure}[t]
\centering
\includegraphics[width=1.0\columnwidth]{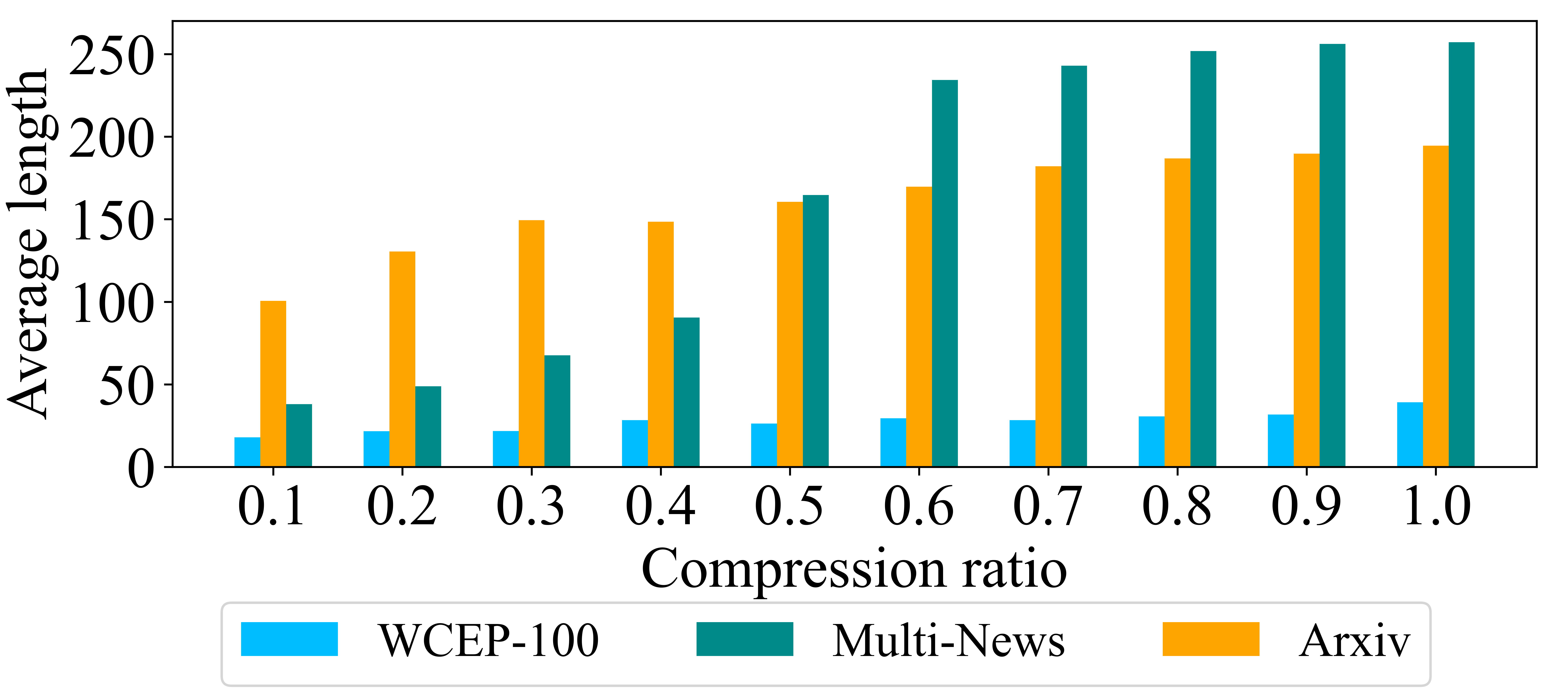} 
\caption{Average lengths of generated summaries for different datasets when the compression ratio $k$ is set to different values.}
\label{fig:k}
\end{figure}

The hyper-parameter $k$ in the heterogeneous graph pooling is to control the proportion of sentence nodes to be retained in the compressed graph. To understand how much $k$ affects the generated summary length, we present average lengths of generated summaries for different datasets when the compression ration $k$ is set to different values in~\figref{k}. Interestingly, we see that larger $k$
generally produces longer summary, and this effect is strongest for \multinews.

\section{Related Work}

\subsection{Abstractive Multi-Document Summarization}

\subsubsection{PLM-Based Models}
Recent PLM-based models have shown strong performance for abstractive text summarization tasks.  These models follow a Transformer-based~\citep{transformer_2017} encoder-decoder architecture. 
For example, general-purpose PLMs such as T5~\citep{t5_2020}, BART~\citep{bart_2020}, and LED~\citep{led_2020} can be fine-tuned for abstractive text summarization.  PEGASUS~\citep{pegasus_2020} is a strong PLM-based model pre-trained with an objective that predicts gap sentences as a pseudo summary. These models can be used for MDS by concatenating the source documents into a single document. 
PRIMERA~\citep{primera_2022} has the same architecture as LED, but is designed for MDS specifically in that it is pre-trained to generate pseudo summaries --- text spans that are automatically extracted based on the entity salience.
%
%
Although these models show impressive performances and can even handle zero-shot cases, they use a flat concatenation of the input documents, which limits their capability in learning the cross-document relationships among different semantic units.

\subsubsection{Graph-Based Models} 
Although graphs are commonly used to boost text summarization \citep{bass_2021, hetertls_2022, hierarchical_hgat_2022}, there are only a handful of models which have been proposed to use graphs to encode the documents in abstractive MDS \citep{graphsum_2020, mgsum_2020, sln_2021, tg_multisum_2021}. 
Most of these models only leverage homogeneous graphs as they do not consider different edge types of graphs.
For example, MGSum~\citep{mgsum_2020} constructs a three-level (i.e., document, sentence, and word levels) hierarchical graph and learns semantics with a multi-level interaction network. GraphSum~\citep{graphsum_2020} constructs a similarity graph over the paragraphs. It learns a graph representation for the paragraphs and uses a  hierarchical graph attention mechanism to guide the summary generation process. %
%
The graphs constructed in these models are in fact homogeneous, in that GraphSum only consider paragraph nodes, and MGSum uses the same edge type to connect the graph nodes. 

\subsection{Graph Neural Networks}

\subsubsection{Graph Modeling}GNNs have yielded strong performance for modeling documents~\citep{gnn4nlp_2021}, e.g., 
to model relationships among text spans for MDS. 
Graph convolutional networks (GCN; \citet{gcn_2016}) and graph attention networks (GAT; \citet{gat_2017}) are two representative GNN models,
which are frequently used in modeling graph-structured data composed of nodes and edges. GAT is based on the attention mechanism~\citep{transformer_2017}, while GCN is based on Laplacian transformation on the adjacency matrix. Another difference between these two is that edge weights of GCNs (i.e., the adjacency matrix) are fixed in training but those of GAT (i.e., the attentions) can be updated, although both of them perform message passing~\citep{message_passing_2017} on graphs. 




\subsubsection{Graph Pooling}
Graph pooling~\citep{graph_pooling_gnn_2022} aggregates node embeddings to obtain compressed graph representations. 
Existing graph pooling methods can be largely grouped into two categories: 
\emph{global pooling} and \emph{hierarchical pooling}. Global pooling generates the graph representation with a mean- or sum-pooling over the node embeddings. This method does not preserve the hierarchical structure of graphs.
Hierarchical pooling, in contrast, considers the graph structure by compressing an input graph into smaller graphs iteratively, through node  clustering~\citep{mincut_pool_2020} or node dropping~\citep{sag_pool_2019}. Our graph compressor follows the idea of the hierarchical pooling, and condenses the graph by removing nodes to generate a small-sized graph. 



\section{Conclusion}

We propose \hgsum, an extended encoder-decoder model that builds on PLMs to incorporate a compressed heterogeneous graph for abstractive multi-document summarization.
\hgsum is novel in that it captures the heterogeneity between words, sentences,  and document units in the constructed graph for source documents, and it also learns to compress the heterogeneous graph by `mimicking' the ground-truth summary graph during training. Experimental results over multiple datasets show that \hgsum outperforms current state-of-the-art MDS systems.



\section{Acknowledgements}
We would like to thank the anonymous reviewers for their great comments. This work is supported by China Scholarship Council (CSC).

\bibliography{aaai23, mreferences}

\end{document}